%% file: anonymous-submission-latex-2023.tex
\documentclass[letterpaper]{article} % DO NOT CHANGE THIS
\usepackage{aaai23} 
\usepackage{times}  % DO NOT CHANGE THIS
\usepackage{helvet}  % DO NOT CHANGE THIS
\usepackage{courier}  % DO NOT CHANGE THIS
\usepackage[hyphens]{url}  % DO NOT CHANGE THIS
\usepackage{graphicx} % DO NOT CHANGE THIS
\usepackage{natbib}  % DO NOT CHANGE THIS AND DO NOT ADD ANY OPTIONS TO IT
\usepackage{caption} % DO NOT CHANGE THIS AND DO NOT ADD ANY OPTIONS TO IT
\usepackage[ruled]{algorithm2e}

\usepackage{bbm}
\usepackage{amsmath,amssymb}
\usepackage{url}
\usepackage{graphicx}
\usepackage{color, colortbl}
\usepackage{algpseudocode}
\usepackage{multirow}
\usepackage[table,xcdraw]{xcolor}
\usepackage{makecell}
\usepackage{bm}
\usepackage{enumitem}
\usepackage{wrapfig}
\usepackage{subfig}

\usepackage[utf8]{inputenc} % allow utf-8 input
\usepackage[T1]{fontenc}    % use 8-bit T1 fonts
\usepackage{url}            % simple URL typesetting
\usepackage{booktabs}       % professional-quality tables
\usepackage{amsfonts}       % blackboard math symbols
\usepackage{nicefrac}       % compact symbols for 1/2, etc.
\usepackage{microtype}      % microtypography

\usepackage[colorlinks]{hyperref}
\hypersetup{
	citecolor = {blue}
}

\def\eqref#1{Eq.~\ref{#1}}
\newcommand{\figref}[1]{Fig.~\ref{#1}}
\newcommand{\tabref}[1]{Table~\ref{#1}}
\newcommand{\secref}[1]{Sec~\ref{#1}}
\usepackage{subfig}
\SetKwComment{Comment}{/* }{ */}

\newcommand{\nsschain}{\textbf{NSS-chain }}
\newcommand{\nsschainxy}{\textbf{NSS-X-chain }}
\newcommand{\nsschainalphay}{\textbf{NSS-$\bm{\alpha}$-chain}}
\newcommand{\nsssum}{\textbf{NSS-sum }}

\makeatletter
\newcommand{\removelatexerror}{\let\@latex@error\@gobble}
\makeatother

\definecolor{Gray}{gray}{0.9} 
 
\usepackage{newfloat}
\usepackage{listings}
\DeclareCaptionStyle{ruled}{labelfont=normalfont,labelsep=colon,strut=off} % DO NOT CHANGE THIS
\title{Neural Spline Search for Quantile Probabilistic Modeling}
\author{
    Ruoxi Sun\textsuperscript{\rm 1}\equalcontrib,
    Chun-Liang Li\textsuperscript{\rm 1}\equalcontrib,
    Sercan \"{O}. Ar{\i}k\textsuperscript{\rm 1},
    Michael W. Dusenberry\textsuperscript{\rm 2},
    Chen-Yu Lee\textsuperscript{\rm 1},
    Tomas Pfister \textsuperscript{\rm 1}
}

\affiliations {
    \textsuperscript{\rm 1}Google Cloud AI
    \textsuperscript{\rm 2}Google Research, Brain Team \\
    \{ruoxis, chunliang, soarik, dusenberrymw, chenyulee, tpfister\}@google.com 
}

\usepackage{bibentry}
\begin{document}

\setlength\parindent{0pt}

\vspace{-1 in}
\maketitle
\vspace{-1 cm}
\begin{abstract}
Accurate estimation of output quantiles is crucial in many use cases, where it is desired to model the range of possibility. Modeling target distribution at arbitrary quantile levels and at arbitrary input attribute levels are important to offer a comprehensive picture of the data, and requires the quantile function to be expressive enough. The quantile function describing the target distribution using quantile levels is critical for quantile regression. Althought various parametric forms for the distributions (that the quantile function specifies) can be adopted, an everlasting problem is selecting the most appropriate one that can properly approximate the data distributions. In this paper, we propose a non-parametric and data-driven approach, Neural Spline Search (NSS), to represent the observed data distribution without parametric assumptions. NSS is flexible and expressive for modeling data distributions by transforming the inputs with a series of monotonic spline regressions guided by symbolic operators. We demonstrate that NSS outperforms previous methods on synthetic, real-world regression and time-series forecasting tasks. 
\end{abstract}
\vspace{-.1 in}
\section{Introduction}
\input{Introduction}
\section{Related Work}

\input{RelatedWork}
\section{Methods}  \label{sec:preliminaries}
\input{Preliminaries}
\input{Methods}
\section{Training}
\input{Train}

\section{Experiments} \label{sec:experiments}
\input{Experiments}

\section{Results} \label{sec:results}
\input{Results}

\bibentry{c:23}.

%\vspace{.2em}
%For the most up to date version of the AAAI reference style, please consult the \textit{AI Magazine} Author Guidelines at \url{https://aaai.org/ojs/index.php/aimagazine/about/submissions#authorGuidelines}

% Use \bibliography{yourbibfile} instead or the References section will not appear in your paper
%\nobibliography{aaai23}
 \bibliography{aaai23} 

\clearpage
\input{Appendix}

%\section{Acknowledgments}
%AAAI is especially grateful to Peter Patel Schneider for his work in implementing the original aaai.sty file, liberally using the ideas of other style hackers, including Barbara Beeton. We also acknowledge with thanks the work of George Ferguson for his guide to using the style and BibTeX files --- which has been incorporated into this document --- and Hans Guesgen, who provided several timely modifications, as well as the many others who have, from time to time, sent in suggestions on improvements to the AAAI style. We are especially grateful to Francisco Cruz, Marc Pujol-Gonzalez, and Mico Loretan for the improvements to the Bib\TeX{} and \LaTeX{} files made in 2020.

%The preparation of the \LaTeX{} and Bib\TeX{} files that implement these instructions was supported by Schlumberger Palo Alto Research, AT\&T Bell Laboratories, Morgan Kaufmann Publishers, The Live Oak Press, LLC, and AAAI Press. Bibliography style changes were added by Sunil Issar. \verb+\+pubnote was added by J. Scott Penberthy. George Ferguson added support for printing the AAAI copyright slug. Additional changes to aaai23.sty and aaai23.bst have been made by Francisco Cruz, Marc Pujol-Gonzalez, and Mico Loretan.

%\bigskip
%\noindent Thank you for reading these instructions carefully. We look forward to receiving your electronic files!

\end{document}

%% file: Introduction.tex
%\section{Introduction}

For many machine learning applications, modeling the prediction intervals (e.g. estimating the ranges all individual predictions observation fall), beyond point estimates, is crucial \citep{salinas2020deepar,  wen2017multi, tagasovska2019single, gasthaus2019probabilistic, pearce2018high}. The prediction intervals can help with decision making for retail sales optimization
\begin{figure}[ht]
%\vspace{-2em}
  \begin{center}
    \includegraphics[width=0.4\textwidth]{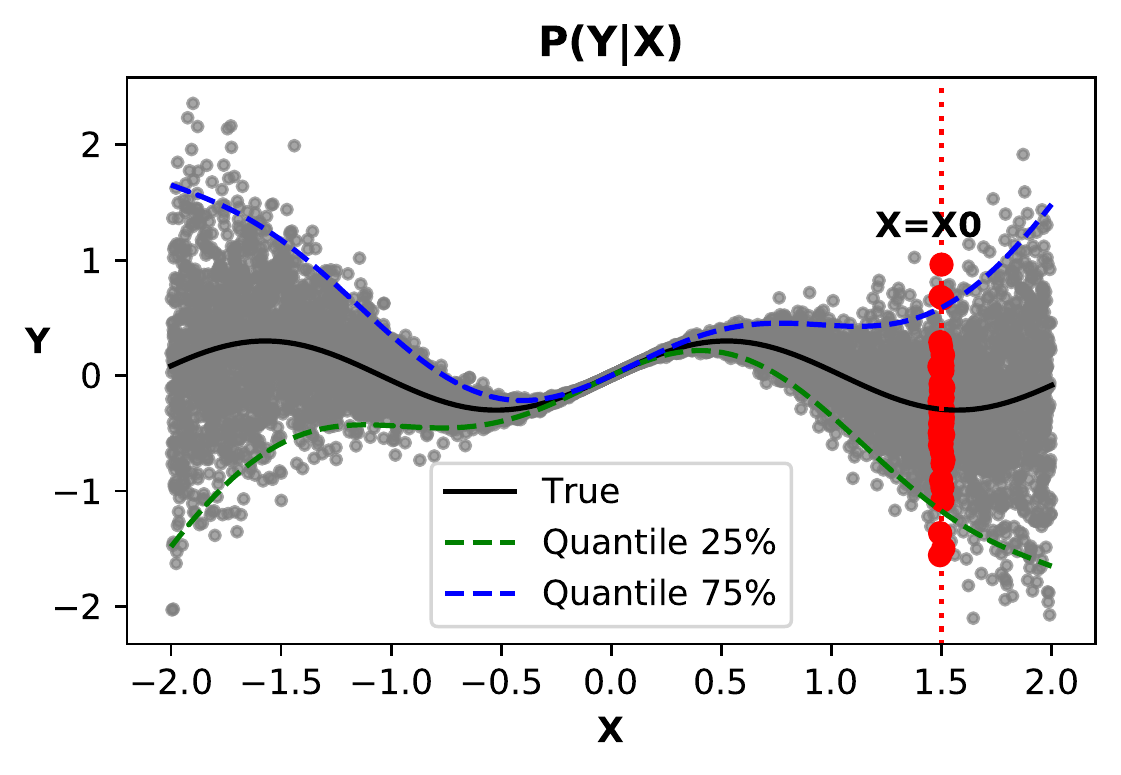}
  \end{center}
\caption{\textbf{Modeling multiple quantiles at different condition-levels with a universal quantile function}. The goal is to model target data distribution {\it y} at any arbitrary quantile level and attribute level {\it X}, using one versatile quantile function. Gray dots are observed data points, while green and blue lines indicate 25\% and 75\% quantile levels.  
%Data distribution Learning data-dependent quantile function for Quantile Regression. 
The data distribution {\it y} varies at different levels of X, say variance of $y$ increases when {\it X} is away from zero. Red dots are data points at  $X=X_0$, $p(Y|X_0)$).    }\label{fig:quantile_regression}
 \vspace{-1.em}
%\end{wrapfigure}
% %Quantile regression models different quantile levels of y with different condition level of X. The quantile function which specifies the data distribution has to work on any arbitrary quantile levels and arbitrary condition level X with any conditional distribution P(y|X). The target value y vary with condition level X with x= 0 with lower variance and .  
\end{figure}
 \citep{simchi2008designing}, medical diagnoses \citep{begoli2019need, mhaskar2017deep, jiang2012calibrating}, information safety \citep{smith2011information}, 
financial investment management \citep{engle1982autoregressive}, robotics and control \citep{buckman2018sample}, autonomous transformation \citep{xu2014motion} and many others.
%\begin{wrapfigure}{R}{0.4\textwidth}

\begin{figure}[ht]
%\vspace{-2em}
  \begin{center}
    \includegraphics[width=0.48\textwidth]{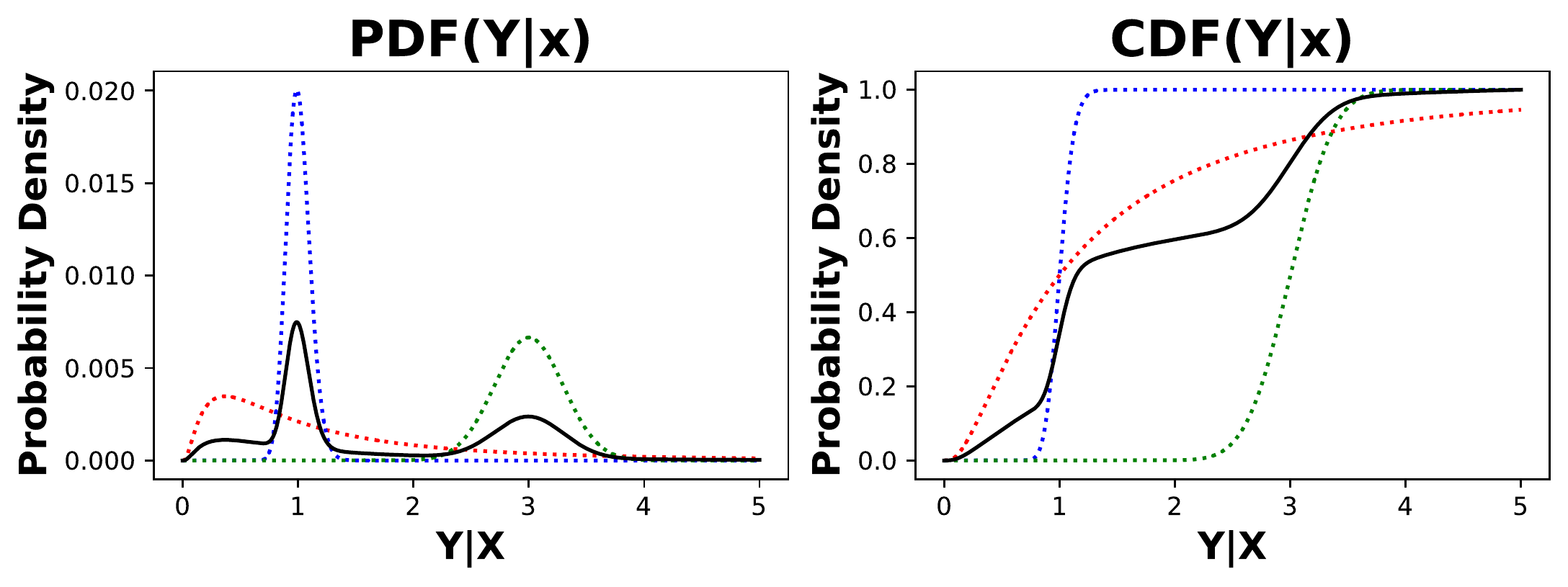}
  \end{center}
\caption{\textbf{An example target distribution with a complex shape, in PDF and CDF space.} Black lines are observed target distributions, in the form of mixture of the other three distributions shown with color. Fitting the black line accurately would be extremely difficult for most of the commonly-used single parametric splines, motivating for the use of learnable spline family composed of multiple splines. %to lead to well fit. %In CDF space, all distribution can be represented as monotonically increasing function. The quantile function is inverse CDF function.  
}\label{fig:pdf_cdf}
 \vspace{-1.em}
%\end{wrapfigure}
% %Quantile regression models different quantile levels of y with different condition level of X. The quantile function which specifies the data distribution has to work on any arbitrary quantile levels and arbitrary condition level X with any conditional distribution P(y|X). The target value y vary with condition level X with x= 0 with lower variance and .  
\end{figure}

%One of the difficulties of classical probabilistic models and those parameterized by neural network for modeling large scale, real world data is the misalignment between the parametric assumptions made on the output distribution and the observed data. 
%Quantifying uncertainty depends on modeling data distribution, which lies in the core of probabilistic modeling on both regression task and time series forecasting. 
%\TODO{Add a connector and motivation} 
To estimate prediction intervals, we would need to estimate different levels of quantiles for the target distribution using quantile regression \cite{koenker2005econometric, waldmann2018quantile}. A real-world challenge is to select the parametric forms of target distributions, which is specified by the quantile function (also known as the inverse CDF function), to properly align with observed data distribution. 
%Selecting the observed data distribution is critical in probabilistic modeling. %data distribution or error distribution is critical for deciding likelihood, 
%For example, conventional linear regression assumes Gaussian errors for the observed data, while its more generalized versions introduce non-Gaussian assumptions, e.g. Poisson or negative binomial likelihood to model sparse count data \citep{nelder1972generalized}. 
%Quantile regression (\figref{fig:quantile_regression}) \cite{koenker2001quantile} aims to predict a quantiles given a condition level (X0). 
Different choices for the target distribution (Gaussian, Poisson, Negative Binomial, Student-t etc.) may yield different quantile predictions, and misalignment of the assumption with the real distribution may hinder the performance of the model. % Although one can make a selection based on prior understanding of the data (e.h. sparse, etc)
%However, the predefined parametetric form  
%Some informative assumptions can be made given prior knowledge about the data (e.g. Poisson priors can be adapted if the observed data is known to contain sparse counts). However, in most cases, 
Therefore, such heuristic or empirical hand-picking based parametric assumptions for the distribution can be sub-optimal. An approach based on learning from the data in an automated way, would be highly desirable, from both foundational and practical perspectives.

%In time series forecasting, a common approach for deep learning is to combine sequential models with a likelihood model to determine the emit density from hidden or latent states to the observations. 
%\citep{salinas2020deepar, wen2017multi, gasthaus2019probabilistic, de2020normalizing, wang2019deep}. While a variety of parametric likelihood (such as Normal, Student-t, Negative Binomial)  is applied to these models to obtain the full probabilistic estimation \citep{nelder1972generalized, salinas2020deepar}, 

%a real-world challenge is determining which likelihood to pick, fitting the data. 

For learnable parametric modeling, one challenge is how to model all quantiles for all input attributes level in a computationally efficient way. %This requires high expressive quantile function that works for . 
First, modeling an any arbitrary quantile, as opposed to a couple of pre-defined quantile levels, offers a more comprehensive view on the target distribution, and provides convenience to use the quantile model (e.g. no need to re-train the model when quantiles at testing are different from the ones at training). Second, real-world data can have complex distributions beyond what simple assumptions can model. \figref{fig:quantile_regression} shows different input attribute $X$ levels have different dependency dynamics with target $y$ level (i.e. the variance of $y$ increases when $X$ apart from $0$). \figref{fig:pdf_cdf} shows that the observed distribution cannot trivially fit well with one single distribution. Therefore, in order to model all quantiles at all $X$, we need a quantile function with a complexity that does not increase significantly with number of input attributes and the number of quantiles. This necessitates a versatile and highly-expressive quantile function. %that can work for arbitrary quantiles and arbitrary input attribute levels. %Doing so require a highly expressive quantile function. 

%Probability distributions can be expressed using quantile functions (i.e. inverse cumulative distribution function)\cite{koenker2001quantile} besides probability density functions. 
%Thanks to the non-parametric nature of quantile space, one can describe probability distribution as a series of anchor knots on the distribution to represent different quantile levels, interpolating between the knots to depict the full probability density. 

There has been many efforts on improving various aspects of quantile regression. \citet{gasthaus2019probabilistic} proposes linear spline interpolation between knots in the inverse CDF space to model the target distribution in time-series forecasting setup. %and incorporate it to hidden state-emission state sequential model target dostronition 
%\citep{wen2017multi, salinas2020deepar} 
This is proposed to avoid the assumption on parametric form of the target distribution. \citet{park2022learning} and \citet{moon2021learning} focus on learning a valid quantile function without quantile crossing (e.g. quantiles violate monotonically increasing property), via special design of the neural network architecture %(based on aggregating non-negative outputs) 
or first-order inequality constraint optimization. %respectively.
Despite being distribution agnostic, these approaches for describing the target distribution (specified by quantile function) are restricted to one function family (e.g. linear spline)% from the input space to the quantile space, 
%with 
, which may limit the expressiveness to represent the target distribution. %\TODO{Update motivation. why high expresstivity is needed}. %Inspired by the success of normalizing flow based methods \citep{rezende2015variational, papamakarios2019normalizing, kingma2016improved, durkan2019neural}, 
In this paper, with the goal of designing an expressive quantile function for various quantiles and input levels, we propose %perform a series of transformations to map the input data to the target distribution guided by a series of symbolic operators. %The quantile function can also be described as a series of transformations on basis transformation (splines) %(\TODO{NOT clear. delete or rephrase}).
a data-driven approach Neural Spline Search (NSS), %NSS offers a flexible and expressive approach for fitting data distribution by 
which transforms the inputs with a series of monotonic spline regressions guided by symbolic operators. 
The contributions of our paper can be summarized as:  %or c-spline-based forecasting \citep{gasthaus2019probabilistic} has shown promising results in modeling complex density distribution. 

\begin{enumerate}
    \item We propose an efficient search space and mechanism to find an expressive quantile function to model the data distribution, avoiding specifying a parametric form of the observed distribution as prior. 
    \item We propose a novel approach to generate an expressive quantile function using a combination of different distributions and operators guided by symbolic operators. %The proposed neural spline search can be generalized to different spline models, including parametric splines. Other spline methods can be integrated into our model as basis splines. 
    \item The proposed method can be incorporated into other tasks (including but not limited to time series forecasting) as their quantile function. 
    \item We demonstrate significant accuracy improvements across numerous regression or time series forecasting tasks. For example, on UCI benchmarks, we show $3.5\%$-$7.0\%$ improvement compared to next best methods. %alternatives. \TODO{be}
\end{enumerate}

\begin{figure*}[h]
%\vskip -0.3in
\centering
\includegraphics[width=1.735\columnwidth]{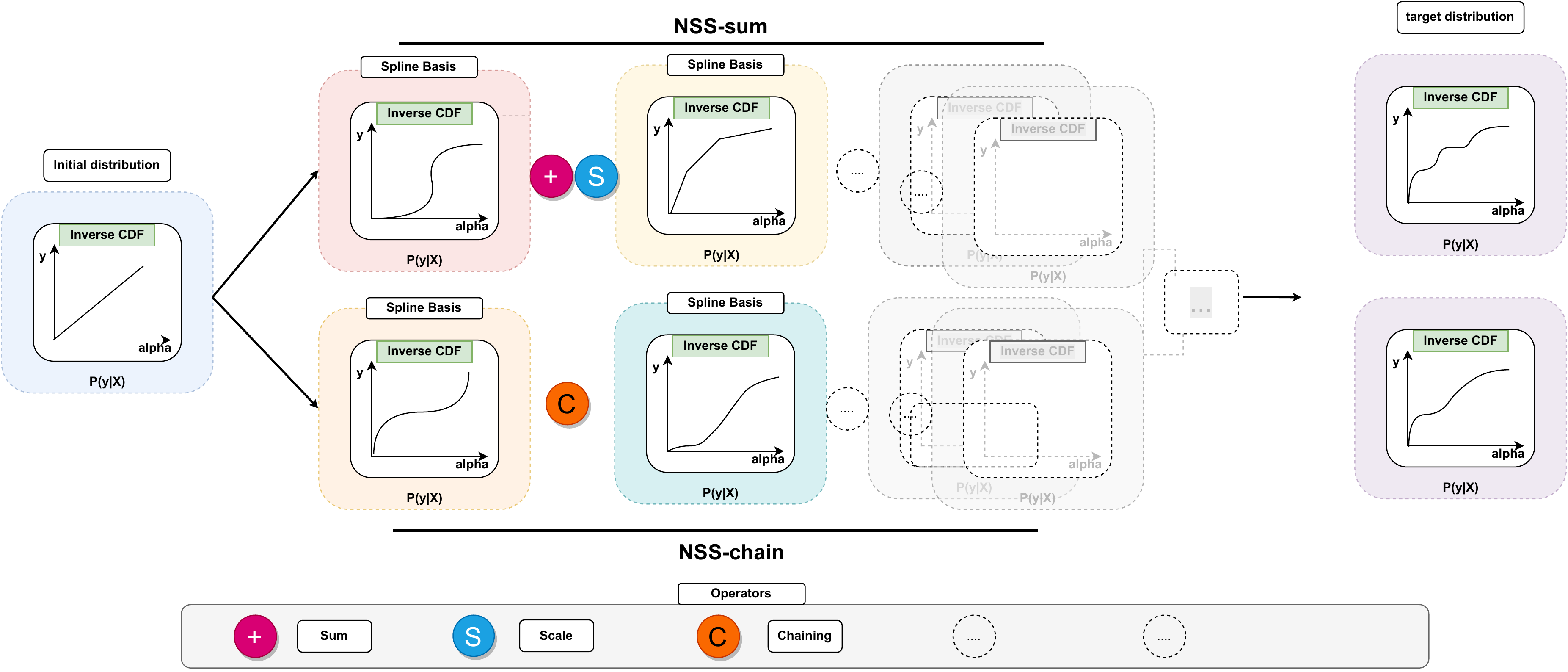}
\caption{\textbf{Overview of Neural Spline Search (NSS)}. %\TODO{Figure caption update; two outcome}
Modeling the target data distribution can be done by learning the quantile function (e.g. inverse CDF), which maps a [0, 1]-variable (quantile) to a target value $y$. 
Unlike parametric methods which specify a distribution family and learn the parameters, NSS can generate the target distribution through a set of transformations on the inverse CDF space (quantile space), where the transformation is guided by a series of operators. 
Here, the bottom gray box shows possible operators (denoted as circles), including but not limited to summation (``+''), scale (``S''), and chaining (``C''). 
The basis splines are shown with color-shaded squares. 
The initial distribution is a uniform distribution, as shown in the leftmost panel (blue shaded), and the target distribution is the rightmost distribution (purple shaded). 
There is no obvious parametric distribution to achieve this transformation. Therefore, NSS is used to search for the suitable transformation through simple operators. 
In the first row of the middle panel, we show operators for {\it NSS-sum}, where the initial uniform distribution is transformed by the red- and the yellow-shaded splines (e.g. c-spline) through sum (``+'') and scale (``S'') operators. 
The second row shows the chaining transformation of the initial distribution, where the orange and cyan splines are used to transform the initial spline. 
The parameters of the splines are learned by a neural network. 
In general, the operators and transformations in NSS are not limited to two splines (we represent them as the gray splines next to the yellow and cyan shaded splines). 
%Descriptions of $X, y, \alpha$ are in \secref{sec:qf}.
}\label{fig:Schematic Picture}
%\vskip -.2 in
\end{figure*} 

%% file: RelatedWork.tex
%Many studies \citep{NguyenYC14, guo2017calibration} have been proposed to improve the model calibration and uncertainty estimation
\textbf{Quantile regression}
is used to estimate the target distribution at different quantile levels. The $\alpha$-quantile estimator is the solution when minimizing quantile loss at level $\alpha$ \citep{koenker1978regression}. Another quantile regression related loss is continuous ranked probability score (CRPS) \citep{gneiting2007strictly}, which is the averaging over all quantile levels, instead of one single quantile. %(as quantile loss %\TODO{NOT self-contained})%\eqref{eq:pinballloss}). 
%shown that the performance of deep neural networks can be poorly-calibrated with conventional training methods. 

%\textbf{Classification problems:} 
%For classification, the posterior probability estimates can be interpreted as confidence estimation, but it calibrates the decision quality poorly \citep{gal2016dropout}. 
%To improve the calibration of classification models, many works have been proposed, such as ensembling methods \citep{lakshminarayanan2017simple}, temperature scaling \citep{laves2019}, distance-based methods\citep{xing2020distancebased} and Bayesian neural networks \citep{pmlr-v119-dusenberry20a}.
%- ADD MORE

%\textbf{Regression problems:} 
%For regression, confidence calibration has been relatively underexplored. 
%For many real-world regression applications like forecasting, quantile regression, along with quantile (pinball loss) is employed to obtain prediction intervals \citep{wen2018multihorizon, lim2021temporal}.
%\citep{Kuleshov2018} proposes extended calibration methods to regression by applying Platt scaling to the cumulative distribution. 

%- ADD MORE
\textbf{Neural network quantile forecasting}. 
To model sequential dependency of time series, several forecasting models propose a hidden state-emission framework (\citep{salinas2020deepar, wen2017multi, gasthaus2019probabilistic, de2020normalizing, wang2019deep}), where the dynamics of hidden states are modeled by auto-regressive recurrent neural works (e.g. LSTM), which takes previous hidden states and current observations as input and outputs current observation. Different from modeling the likelihood with parametric distributions (e.g. Gaussian~\cite{salinas2020deepar}), emission models for quantile estimation is to learn the parameters of quantile function. The overall framework is optimized by employing a quantile \cite{wen2017multi} or CRPS \citep{gasthaus2019probabilistic} loss.  %The loss can be either quantile loss \cite{wen2017multi}, which optimizing against single quantile level, or continuous ranked probablisity score (CRPS), which is overtimizing over an integration of all quantile levels)\citep{gasthaus2019probabilistic}. 

\textbf{Symbolic regression} has shown great success in many fields, including program synthesis \citep{parisotto2016neuro}, mathematical expressions extraction \citep{ cranmer2020discovering}, physics-based learning \citep{li2019neural, petersen2019deep}. 
As the search space is enormous and scaled exponentially with the length of operators, symbolic regression rule operators are usually set to be a small number and are learned by Monte Carlo Tree Search guided evolutionary strategies \citep{li2019neural} or reinforcement learning \citep{petersen2019deep}. 
%\section{Related Work}

%% file: Preliminaries.tex
%\section{Preliminaries}  \label{sec:preliminaries}
\subsection{Learning quantile function in quantile regression}\label{sec:qf}
%\TODO{described c-spline/p-spline in details}
%Regression is to estimate conditional mean of the target variable given certain level of input attributes. 

%Quantile regression estimates different conditional quantile levels of the target variable given a certain level of input attributes, as opposed to regression, which estimates the conditional mean of the target variable. 
Let the input data attributes $X$ and the target variable $y$ are jointly distributed as $p(X, y)$. The conditional cumulative distribution function (CDF) is $F(Y=y|X) = P(Y \leq y | X)$. The quantile function, which is also called the inverse CDF function, takes quantile level as inputs and returns a threshold value $Y$ below which random draws from the given CDF would fall quantile percent of the time. Specifically, 
%describes the target distribution with input as quantile levels and output as target random variable whose CDF value is the quantile, is critical for quantile regression. the quantile function Q returns a threshold value x below which random draws from the given c.d.f. would fall p percent of the time.
the $\alpha$-th quantile function of $y|X=x$ is denoted as:
\begin{equation}
q(\alpha, x) = F^{-1}_{y|X=x}(\alpha) = \inf\{y: F(y|X=x) \geq \alpha\}
\end{equation}
%Given a certain level of input attributes $X$, 
Here we can think the quantile function is to perform a transformation on a uniform-distributed random variable $\alpha \sim U(0, 1)$ to the target distribution $p(y|X)$. Quantile function is able to fully specify a distribution. So specifying the quantile function is describing the target distribution $p(y|X)$. \\
%\TODO{Add figure }
%\subsection{Conditional Quantile Regression}

Quantile regression estimates different conditional quantile levels of the target variable given a certain level of input attributes, as opposed to regression, which estimates the conditional mean of the target variable. In quantile regression, a particular quantile level $\alpha$ of the conditional distribution of $y$ given $X=x$, $q(\alpha, x)$ %= $F^{-1}_{y|X=x}(\alpha)$ 
is estimated by minimizing the {\it pinball loss $\rho$} (or quantile loss), as the the quantile function $q$ is shown to be the minimizer of the expected pinball loss \citep{koenker1978regression}: 
\begin{equation}
\rho^{\alpha} (y, q) = (y - q)(\alpha - \mathbbm{1}{(y < q)}), \label{eq:pinballloss}
\end{equation}

\begin{equation}
q(\alpha, x) = \arg\min_q \mathbb{E}_y[\rho^{\alpha}(y,q)]. \label{eq:expected-pinballloss}
\end{equation}
where $\mathbbm{1}$ is the indicator function. %, $\alpha$ is the quantile level and $q$ is the predicted $\alpha$-th quantile of $y$ at the input attribute level $x$. 
%In order to estimate $q$, we parameterize the quantile function as $q_{\theta}$, where $\theta$ are learnable model parameters. The quantile function takes $X$ and quantile level $\alpha$ as inputs.
%During training, the best-fit parameters are found by optimizing over with the empirical mean of CRPS over $N$ data points ($X_i$, $y_i$), we get:
%\begin{equation}
%\theta^* = \arg\min_{\theta} \frac{1}{N} \sum_{i=1}^N %\rho^{\alpha}(y_i, q_{\theta}(X_i, \alpha)).
%\end{equation}
%Note that $\theta$ are shared between different conditional vales ($X$) \TODO{inpout?} and quantile levels ($\alpha$). 
One shortcoming of pinball loss is only measuring the loss at a single quantile level, which hinders the estimated $q$ for a global picture of the distribution (i.e. other $\alpha$ levels). On contrast, the {\it continuous ranked probability score} (CRPS) considers all quantile levels by integrating the pinball loss over $\alpha = [0, 1]$ \citep{matheson1976scoring,gneiting2007strictly}. 
\begin{equation}
    \text{CRPS}(y, q) = \int_0^1 2\rho^{\alpha}(y, q)d \alpha \label{eq:crps}
\end{equation}
% The advantage of optimizing CRPS compared to quantile loss is that CRPS takes into account of all quantile levels, not only a single quantile.
As a proper scoring rule \cite{gneiting2007strictly},  CRPS is minimized when the quantile function is $q = F$. That is,
\begin{equation}
F_y^{-1} = \arg\min_q \mathbb{E}_y[\text{CRPS}(y, q)].\label{eq:crps-theory}
\end{equation}
Please refer \citep{koenker2005econometric} for detailed proof. %Finally, during training, we fit parameters by optimizing over with the empirical mean of CRPS over $N$ data points:
%\begin{equation}
%\theta^* = \arg\min_{\theta} 1/N \sum_{i=1}^N %\mathbb{E}_y[\text{CRPS}(y, q_{\theta}(X_i, \alpha))].
%\end{equation}

%For more details on the proof, please see \citep{gasthaus2019probabilistic, koenker2005econometric}. %The key to probabilistic forecasting is to estimate the quantile function, which is a monotonically-increasing function that maps $\alpha \in [0, 1]$ to the target $y$ values. %, the probability of being smaller than $\alpha$. 
%Any function $q: [0, 1] \rightarrow \mathbb{R}$ that is monotonically increasing and left continuous can serve as a quantile function, which 
%Quantile function also uniquely defines a probability distribution $F$, where $q = {F^{-1}}$. We choose the quantile (inverse CDF) space to perform the following transformation, because (1) the constraints of being bounded in $[0, 1]$ and being monotonically increasing are easier to impose compared to the constraint of PDF, e.g. integrate to one; and (2) it is easier to obtain prediction level and sample from the distribution (e.g. sample from $u \sim \text{Uniform}(0, 1)$ and map to $y$ space using $y=F^{-1}(u)$). 
%Can we define a transformation space with basis as monotonic splines upon which we transform uniform distribution to can generate expressive quantile functions to expressively describe input data distribution? Without loss of generality, we select non-parametric splines c-spline and p-spline as our basis in spline space. We start by defining them: 
\subsection{Improving the expressiveness of quantile function}
%\TODO{[Add connection paragraph] introduce the methods}
% As mentioned in previous section,  
% learning the quantile function $q$ is critical in quantile regression. 
% Quantile functions 
% map quantile levels $\alpha \in [0, 1]$ to the target $y$ values. It also uniquely defines a CDF distribution $F$, as $q$ is inverse CDF function $  {F^{-1}}$. The critical question is how to represent the quantile function to better achieve the mapping from [0, 1] to the target distribution. We denote applying a distribution as a transformation of the distribution. 
\figref{fig:pdf_cdf} demonstrate the need of an expressive quantile function for modeling target distribution. % that if we focused on transformation in CDF space. We denote a To this end, 
Inspired from neural architecture search (NAS) \cite{elsken2019neural}, we propose an approach to search for the suitable combination of distributions. %to specify an expressive quantile function. 
The search is over different operations and basis distributions. %in inverse CDF space. %It is inspired from neural architecture search (NAS) \cite{elsken2019neural} for searching optimal architecture building blocks or layers. %In NSS, on the other hand, the search space is over distributions and operations are on distributions. %Some foundation behind doing so is the combination of monotonic increasing splines or stack of splines on each other yields monotonic increasing splines, so combining different splines preserves monotonic increasing property. 
%We started by introducing our 
%Based on parametric distribution, we also introducing spline-based distributions \emph{c-spline} and \emph{p-spline}% \TODO{union them}
%, which are non-parametric.  We use spline-based distibutions 
%However, we note that the proposed idea can be extended to other basis functions. Our basis distribution can include parametric, or non-parametric. For parametric, if we know some prior knowledge the target distribution is likely to have some distribution (e.g. Normal noise), we would better to include the parametric distribution (it is hard to learn normal CDF from scratch using spline based algorithm)  
We first introduce parametrization of quantile function, and the two non-parametric spline-based distributions. 
\subsubsection{Parameterizing quantile functions}
We propose to parameterize the quantile function $q_{\theta}(\alpha, x)$ using a deep neural network with parameters $\theta$. The quantile function is aimed to be accurate for any quantile levels $\alpha$ and input attributes level $X=x$. %$\theta$ are the weights of %as a neural networks with parameters $\theta$ are learnable model parameters. The quantile function takes $X$ and quantile level $\alpha$ as inputs.
%During training, the best-fit parameters are found by optimizing over with the empirical mean of CRPS over $N$ data points ($X_i$, $y_i$), we get:
%\begin{equation}
%\theta^* = \arg\min_{\theta} \frac{1}{N} \sum_{i=1}^N %\rho^{\alpha}(y_i, q_{\theta}(X_i, \alpha)).
%\end{equation}
%Note that $\theta$ are shared between different input attributes ($X$) and quantile levels ($\alpha$). 
$X$ is high dimensional in real data, not as the one dimensional in the toy examples in \figref{fig:quantile_regression} and \figref{fig:pdf_cdf}. 
\subsubsection{C-spline distribution}
%\TODO{Polish, add details}
 The c-spline ($y^{\alpha} = q^{csplie}_{\theta}(\alpha, x)$) describes the CDF  (\figref{fig:pdf_cdf}, Right Panel) of a probability distribution $F_{y|X}$ by setting $K$ anchor points (denoted as knots) on the CDF curve and performing linear interpolation to fill in the gap between the knots. Specifically, the knots split CDF curve into bins and c-spline learns the width $w_i$ and height $h_i$ of bins %for CDF. %Specifically, we anchor $K$ knots on CDF curve and perform linear interpolation between them. 
 %The heights $h_i$ and widths $w_i$ of the knots are learned 
 by neural networks $\text{NN}$ that depend on the input attributes level $X=x$. 
\begin{align*}
    &\{w_i, h_i\}^K = \text{NN}_{\theta}(x) \\
    &y^{\alpha} = r(\{w_i, h_i\}^K, \alpha) \quad \forall \alpha \in [0: 1]
\end{align*}
where $h_i$ and $w_i$ are non-negative delta values imposed by non-negative activation (i.e. Relu or Sigmoid), and the location of each bin (e.g. Y|X) is $L_i = \sum_{k=0}^i w_k$ and quantile level $\alpha_i = \sum_{k=0}^i h_k$. The accumulation sum design is to ensure that quantile function is monotically increasing and there is no quantile crossing. $r$ is a function to convert knots to output of quantile function: for quantile level $\alpha_i$ that is on the knots, we can directly read from $l_i$ , for quantile levels that are off the knots, quantile values can be computed through linear algebra operations on the two nearby knots
%\vspace*{ 2\baselineskip}
%\resizebox{\hsize}{!}{
%\[
    $r(\alpha)= 
\begin{cases}
    l_i + \frac{(\alpha - \alpha_i)(l_j - l_i)}{\alpha_j - \alpha_i}, & \text{if } \alpha_i \leq \alpha \leq \alpha_j  \quad  0 \leq i, j \leq  K\\
    l_k,              & \text{if } h_k = \alpha 
\end{cases}$
%\] 
%}

%f =  when $$
%where $m$ is neural networks that takes input 

%Note that although c-spline has its own parameters such as $\theta$, but it is not para
%$\omega$ contain $K$ notes information, including height and width. $\omega$  is used to reconstruct the entire distribution 
    
% The c-spline describes the CDF of a probability distribution $F_{y|X}$. It sets $K$ anchor points (denoted as knots) on the distribution. By linearly connecting the knots, c-spline obtains the full distribution. The statistics of the K-knots are $\omega = \{y_i, h_i\}^K$ consisting of their locations and heights, are the major parameters to be learned for c-spline. As the shape of c-spline depends on condition level $X$, $\omega$ depend on $X$: $\omega(X) = m_\theta(X)$, where $m_{\theta}$ is some neural networks with weights $\theta$. 
% %serving as parameters of c-spline.  A learnable model  $m_{\theta}$ can be used to model c-spline parameter $\omega$ with $X$ as input to depict the input level specific $y$ distribution $\omega(x) = m_\theta(X)$. 
% Therefore, the c-spline regression is defined using quantile function: $q_{\theta}^{(c)}: X \in \mathbb{R}^d, \alpha \in [0,1] \rightarrow y \in \mathbb{R}^1$, where any level of quantile $\alpha$ can be computed through linear algebra operations on the defined knots. The K knots are implemented by non-negative incremental between the two consequent knots (the nonnegativity can be satisfied by passing through a nonlinearity like a Sigmoid or ReLU), therefore no quantile crossing happens. %\TODO{why C-spline is not enough} 
%\vspace*{-2\baselineskip}

\subsubsection{P-spline distribution}
%\TODO{[Add connection paragraph] Difference from c-spline} \TODO{Polish} 
The difference between p-spline from c-spline is having anchor knots in PDF space, instead of CDF space. Similarly with C-spline, P-spline also perform linear interpolation over knots, and the quantile level is achieved by integration over pdf via polynomial operations. 

%% file: Methods.tex
\section{Neural Spline Search (NSS)} \label{sec:methods}
%\vspace*{-2\baselineskip}
%Fig. 2 (Right Panel) demonstrate the need of an expressivequantile function for modeling distribution. Therefore, wepropose an approach to search for the suitable combinationof different distributionS to obtain an expressive quantilefunction. The combination denotes operations over distribu-tions. The basis distributions are any distribution in quantilespace (inverse CDF). It is inspired from neural architecturesearch (NAS) (Elsken, Metzen, and Hutter 2019) for search-ing optimal architecture building blocks or layers.
%\secref{sec:methods}
%\nameref{sec:methods}
%\cref{sec:methods}
%\section{Neural Spline Search (NSS)} \label{sec:methods} %\TODO{To polish! Add details}

%\TODO{add introduction and connecting sentences }
%\section{Methods} \label{sec:methods}

 %   \end{minipage} %{0.5\textwidth}
%\TODO{[Add connection paragraph] Why NSS? What problem it solves?} 
We describe our proposed method, Neural Spline Search (NSS), %, to improve fitting the target data distribution. %by effectively searching in the space of discrete symbolic operators to represent the output transformation. 
which is overviewed in \figref{fig:Schematic Picture}. 
%We phrase the proposed algorithm Neural Spline Search (NSS) in the language of symbolic regression, which searches a series of discrete symbolic rules (operators) to fit the target data distribution. 
%Symbolic regression has shown great success in many fields, including program synthesis \citep{parisotto2016neuro}, mathematical expressions extraction \citep{ cranmer2020discovering}, physics rule learning \citep{li2019neural, petersen2019deep}. As the search space is enormous and scaled exponentially with the length of operators, symbolic regression rule operators are usually set to be a small number and are learned by Monte Carlo Tree Search guided evolutionary strategies \citep{li2019neural} or reinforcement learning \citep{petersen2019deep}. 
Similar to symbolic regression \cite{parisotto2016neuro, li2019neural}, NSS effectively searches in the space of discrete symbolic operators and distribution space for a candidate that can better fit the target data distribution. %Here, we define {\it transformations} are for a variable's original distribution using the quantile function (associated with the transform distribution).   
Specifically, let $T(O, S, k)$ denote the space of all transformations, via operators $O$ on all distribution $S$ with a maximum sequence length $k$. NSS aims to find the function $f(x)$ selecting operators and distributions in the space $T$ such that \{$f(x) \in T(O, S, k):
\ell(f(x), x_{train}) \leq \delta$ \}, where $ \ell$ denotes loss function CRPS,  $x_{train}$ is training data %, e.g. root mean square error (RMSE) \TODO{Change to CRPS?} 
and $\delta$ is the acceptance threshold. 
%\begin{wrapfigure}{r}{0.6\textwidth}
%  \begin{center}
    %\includegraphics[width=0.48\textwidth]{birds}
    %\includegraphics[width=.58\columnwidth]{Figures/synthesized.pdf}
 % \end{center}
%  \caption{Birds}
%\end{wrapfigure}
Given  the  large  search space composed of combinations of numerous splines and operators,  %constraints on the unknown rules would be needed to make the search more efficient. Here, quantile space transformation requires to be monotonically increasing and from the range [0, 1]. %These two constrains largely reduce our search choices, therefore we select our basis spline as: p-spline and c-spline. 
%As the search space for splines and the transformation operators is unlimited, 
%For simplicity, 
we restrict to use spline-based distribution as the basis distribution, and limit the operator search space to summation and chaining operations upon the transformation basis spline regressions. Note that this work can be easily extend to other operations and distributions, which we leave to future work. 
%We note that our work can be easily extended to other operators or splines, including parametric distribution transformation (i.e. Gaussian, Cauchy, or Negative Binomial). 
%Practically, due to the prohibitive computational burden of optimizing over splines, we do not perform an exhaustive %explicit 
%search over all possible $T$. 
%Instead, we propose 
We describe the following NSS transformations as they are observed to work well consistently across different datasets: %We present them in the following subsections.  %: c-spline and p-spline. 
%Our work can be easily extended to other operators or other splines, including parametric distribution transformation (e.g. Gaussian or Cauchy distribution). However, these explorations are beyond the scope of this work and we will leave it to future work. 
%We describe two simple transformations $f$ that work well in practice: 
NSS with summation (NSS-sum) and NSS with chaining (NSS-chain). Algorithm \ref{alg:Neural Spline Search} and \figref{fig:cartoon}(b) %\figref{fig:synbolic}. %For simplicity, we demonstrated using two splines.  
%\begin{figure}[ht]
%\centering
%\includegraphics[width=.6\columnwidth]{Figures/syn}
%\caption{\textbf{Symbolic operators for Neural Spline Search-sum (NSS-sum) and Neural Spline Search-chain}. C indicates ``chaining''. }\label{fig:synbolic}
%\end{figure} 
%\vspace*{-2\baselineskip}

\subsection{NSS-sum}   
% \vspace*{-2\baselineskip}
%One NSS design is NSS-sum. 
NSS-sum performs transformations using the scale and summation operators. 
We represent this scenario with two splines: Spline 1: c-spline and Spline 2: p-spline, and two operators: scale $O1: O(a) = \lambda a$ and summation $O2: O(a, b): a + b$; therefore, the overall transformation is (Spline 1-Operator 1) - (Spline 2-Operator 2), which yields: $f$ = c-spline + $\lambda$ p-spline. Essentially, NSS-sum performs weighted sum of different splines. The motivation behind is that c-spline with fewer parameters can be more robust against overfitting, whereas p-spline increases the expressiveness of the splines. 
%\begin{figure*}[ht]
%\centering
%\includegraphics[width=.5\columnwidth]{Figures/quantile_NSS_sum_chain.drawio_1}
%\caption{ \textbf{Illustration of NSS-chain methods}. The diagram demonstrates the chaining loop of the NSS-chain. \textbf{Left:} $\alpha y$ loop, where the output $y$ of the spline, after re-scaled to [0, 1], is re-inputted to the quantile spline as quantile level $\alpha$. \textbf{Right:} $Xy$ loop, where the output $y$ is instead re-inputted to the quantile spline as $X$. The parameters of condition-specific splines are generated by the transformation of the condition value $X$ using some MLP neural network. Description of $X, y, \alpha$ is in \secref{sec:qf}.}\label{fig:NSS-chain methods}
%\end{figure*} 
%\begin{figure}
%\centering
%\begin{minipage}{.5\textwidth}
%  \centering
%  \includegraphics[width=.5\linewidth]{Figures/quantile_NSS_sum_chain.drawio_1}
%  \captionof{figure}{A figure}
%  \label{fig:test1}
%\end{minipage}%
%\begin{minipage}{.5\textwidth}
%  \centering
%  \includegraphics[width=.5\linewidth]{Figures/syn}
%  \captionof{figure}{Another figure}
%  \label{fig:test2}
%\end{minipage}
%\end{figure}
%\vspace(-1in)
%\vskip -23.15in
%\vspace*{-2\baselineskip}

\subsection{NSS-chain} 
\begin{figure}[t]
% \vskip +0.2in
\centering
\includegraphics[width=1.0\columnwidth]{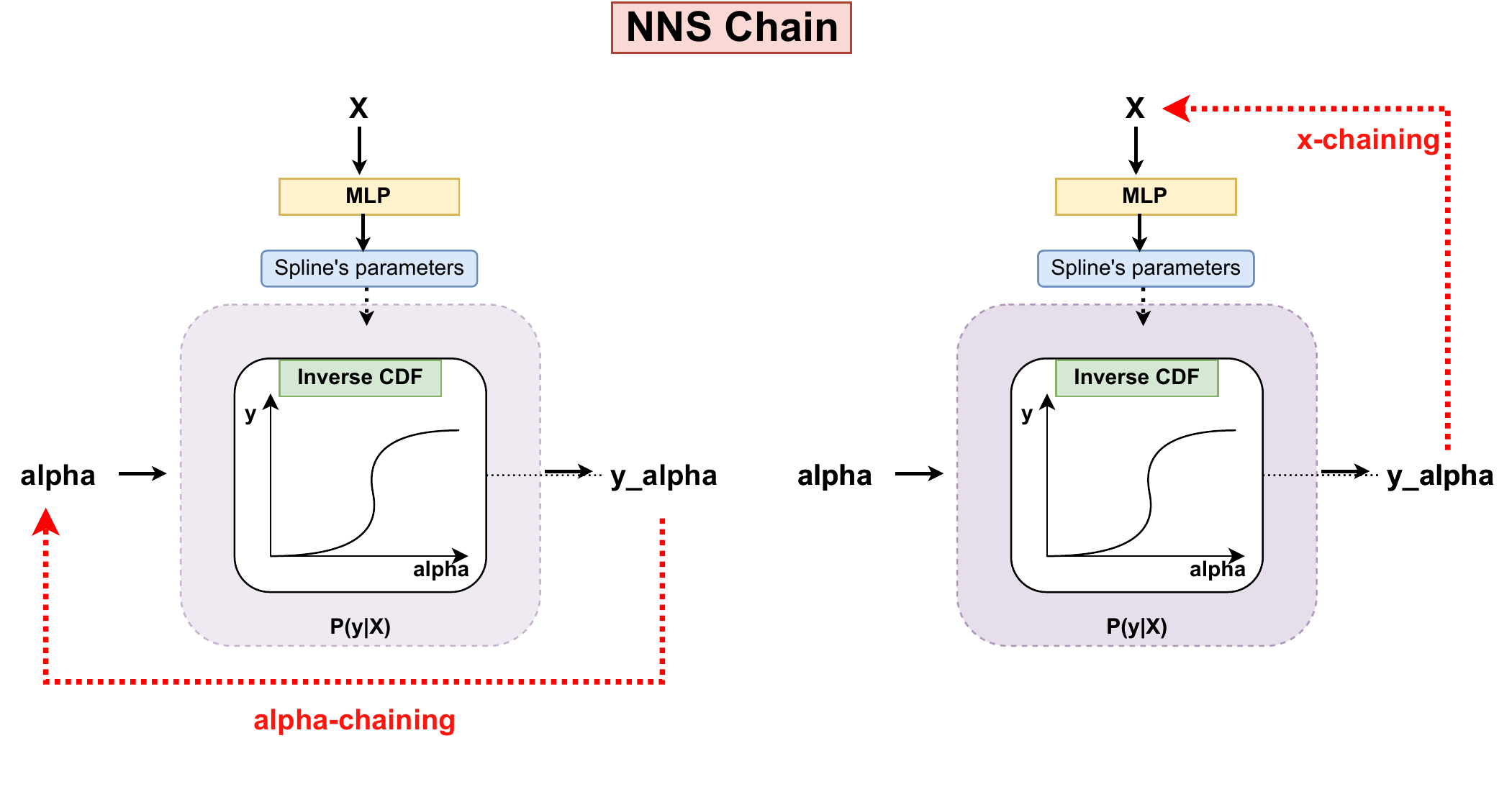}
\caption{(a) \textbf{Illustration of NSS-chain methods}. The diagram demonstrates chaining for NSS-chain. \textbf{Left:} $\alpha$-chaining. The output $y$ of the spline, after re-scaling to [0, 1], is re-inputted to the quantile spline at quantile level $\alpha$. \textbf{Right:} $X$-chaining. The output $y$ is instead re-inputted to the quantile spline as $X$. Both rely on input attributes $X$. %The parameters of condition-specific splines are generated by the transformation of the condition value $X$. %zDescription of $X, y, \alpha$ are provided in \secref{sec:qf}.
}\label{fig:cartoon}
\vskip -0.05in
\end{figure}

%\vskip -.15in
Another proposed NSS design is NSS-chain. We focus on the chaining operator due to its expressiveness. This design is inspired by the success of normalizing flow \citep{rezende2015variational}, where a sequence of bijector transforms is utilized to transform distributions. %on the input distribution to map it to the complex target distribution. %Normalizing flow is considered as expressive for modeling complex probability distribution. 
Different from normalizing flow which has practical applicability challenges, NSS-chain only requires the forward pass of the transformation, not the inverse as normalizing flow does. This significantly reduces the computational complexity and broadens the feasibility of transformations.    
%For simplicity and demonstration purpose, we present a two-steps of transformation chaining.  %NAS-chain OPERATORS we investigated are the following two:  
As mentioned, quantile function takes input attributes level ($X$) to predict the target value ($y$) at quantile level ($\alpha$). 
\begin{equation}
     y = q_\theta(X, \alpha), \label{eq:quantile} 
\end{equation}
where $X \in R^m$ and $\alpha \in [0, 1]$. %The chaining represented as 
We present two designs to chain different transformations (see \figref{fig:cartoon} (a)). We note that chaining of transformation is not limited to the two designs.
%\vspace*{+5\baselineskip}
     \begin{minipage}{0.5\textwidth}
     \vskip 0.23in
     \removelatexerror% Nullify \@latex@error
 \begin{algorithm}[H] %[ht]
\caption{Neural Spline Search} \label{alg:Neural Spline Search}
\textbf{Operators} = \{+, $\times$, Scale, Chain, ...\} \\ 
\textbf{Splines} = \{c-spline, p-spline, Gaussian, Cauchy ...\} \\
%\textbf{Transform}   \\
\KwData{Quantile level $\alpha \in [0, 1]$, $N$ data points $\{X \in \mathbb{R}^{d},  y \in \mathbb{R}^1  \}_N, d \geq 1$, with chain depth $k$. Transform indicates the transformation using the input spline $S_{\theta}$ and operator $O$.}
%\KwInput{$n \geq 0$}
\KwResult{$p(y|X)$ and $F^{-1}_{y|X}(\alpha)$}
%$y \gets 1$\;
%$X \gets x$\;
$k \gets 1$\;
\While{$k \leq K$}{
%   \eIf{$N$ is even}{
%     $X \gets X \times X$\;
%     $N \gets \frac{N}{2}$ \Comment*[r]{This is a comment}
%   }{\If{$N$ is odd}{
%       $y \gets y \times X$\;
%       $N \gets N - 1$\;
%     }
    
%   }
   \textbf{Select} $O = \{O_i\}_{no} \in$ Operators  \; 
   \textbf{Select} $S = \{S_j\}_{ns} \in$ Splines  \; 
   $\theta \gets \textbf{MLP}(X)$ \; 
   $ y_{pred} \gets \text{Transform} (S_{\theta}, O, \alpha)$\;  %\Comment*[r]{Normalize $y$ to [0, 1]}
   
   \eIf{$\alpha$ \nsschain}{
   %\ELSIF{$\alpha y$ \nsschain}{
   \textbf{Normalize} $y_{pred}$ to $[0, 1]$ as $y'_{pred}$ \;   %\Comment*[r]{sigmoid or scale$^{[1]}$}
    $\alpha \gets y'_{pred}$\;
    %$N \gets \frac{N}{2}$ %\Comment*[r]{This is a comment}
  }{%\If{$Xy$ \nsschain}{
      %$y \gets y \times X$\;
     $X \gets Y$  \algorithmiccomment{ if X-\nsschain} \;
      
   % }
    
  }
  
   $k \gets k + 1$\;%https://braintex.goog/project/615d02bec26c06007e99a540
}
 \vskip +0.05in
\end{algorithm} 

\end{minipage}

\begin{itemize}[{leftmargin=*}] %[noitemsep,topsep=1pt,parsep=1pt,partopsep=0pt, leftmargin=*]
\item \textbf{$\alpha$-chaining} \\
The $\alpha$-chaining is when we consider the condition level ($X$) unchanged during the chain of transformation, and the output of each transformation is a scaled version of quantile level for the next transformation. %, we have \textbf{$\alpha$} chainning.   %The quantile function $q$ takes condition level $X$ and quantile level $\alpha$ as input and outputs a scalar value as the estimated target value $y$. 
%$g_{\theta}^{(p)}: X \in \mathbb{R}^m, \alpha \in [0,1] \rightarrow y \in \mathbb{R}^1$
%When we consider the condition level ($X$) is unchanged during chain of transformation, and the output of the transformation is to learn a scaled version of quantile level for the next transformation. 
In particular, after each transformation, we normalize the output $y$ to be in the range $[0, 1]$, and then the normalized output is re-input as the new $\alpha$ to the next transformation. This is repeated until the maximum depth is reached. This design is more similar with normalizing flow methods.  %we exhaust the SELECTION collections. 
\begin{equation}
 y = q_{\theta_K}( X, ...f_n( q_{\theta_2}(X, f_n(q_{\theta_1}(X, \alpha)))))
\end{equation}
$\theta_k$ for $k$=1,2,..$K$ are parameters for different splines in K-length chain. $f_n$ is the normalization function. 
\item \textbf{$X$-chaining} \\
%\subsubsection{NAS-chain p-spline (c-spline())} 
$X$-chaining is when we consider quantile level $\alpha$ level is unchanged during chaining, as each transformation learns a suitable condition level (or feature) for next iteration. Similarly with $\alpha$-chaining in the iterative manner, except that the output $y$ of each transformation is projected to generate $X$ for the next iteration of \eqref{eq:quantile}. 
\begin{equation}
 y = q_{\theta_K}( ... q_{\theta_2}(q_{\theta_1}(X, \alpha), \alpha), \alpha)
\end{equation}
The advantage of this approach, compared tp $\alpha$-chaining, is that we keep quantile levels $\alpha$ unchanged, and re-normalizing output is not needed. 
\end{itemize}
%Details see %\algref{alg:Neural Spline Search}. 
%\TODO{[Address R1's comments]
%"The actual technique proposed is also not well motivated. How chaining different quantile functions increases modelling capability? Because a more complicated quantile function can in fact be modeled by a spline by choosing a large number of knots/pieces in the unit interval [0, 1]? Isn’t this modeling flexibility the main idea behind splines?
%Moreover, chaining does not even make sense syntactically as the domain and range of the components in this chain do not match. The domain of the first component in the chain is the unit interval and the range is same as that of target value. They re-scale it to [0, 1] in order to feed this as input to the second component. This is definitely not a good design of transformation chain."}

\textbf{Remarks on NSS: }. 
\textbf{(1) why a simple  spline-based  algorithm, e.g. C-spline, is not enough?} Although in theory spline-based algorithms can represent any arbitrary distributions with sufficiently high number of knots $K$, in practice, we find a large $K$ often lead to unstable training, as also studied in \cite{park2022learning}. In contrast, we find the combination (combined or chained) over a relatively restricted splines are more robust in capturing the overall of the target distribution
%The application of multiple splines combined or chained is superior than increasing the number of knots, as the latter gives too many degrees of freedom and training becomes more challenging \cite{park2022learning}. 
%As an intuitive example, it is not trivial to reconstruct a Gaussian distribution using infinite number of knots. 
\textbf{(2) Include both spline-based distribution and classic parametric distribution}
In addition to spline-based distribution, we also encourage incorporating parametric distribution (e.g. Gaussian) as basis distribution for NSS, especially when prior knowledge (say Gaussian noise) is available. Because, it is challenging for spline based methods to reconstruct Gaussian distribution even with infinite number of knots; and , the benefits of combining the two are the parametric distribution offers advantage of classic statistics and robust to noise, and the non-parametric spline offers flexibility. %Second, %Though the approach is not non-parametric any mo%Therefore a balance between number of knots and demonstration of usefulness distribution is important. Second, 
%\vspace{

%% file: Train.tex
%\section{Training}
%\vspace{-1em}

\begin{table*}[ht]
\centering
\resizebox{0.6\textwidth}{!}{\begin{tabular}{l|c|c|c|c|c|c}
%\hline
\Xhline{2\arrayrulewidth}
\textbf{Methods} &
 % \textbf{Measures} &
  \textbf{Boston}&
  \textbf{Concrete} &
  \textbf{kin8nm} &
 \textbf{Power} &
  \textbf{Protein} &
  \textbf{Wine} \\
  %\hline
\Xhline{2\arrayrulewidth}
\textbf{Gaussian}  &
%  MAE &
  0.0754 &
  0.0564 &
  0.048 &
  0.0449 &
  0.2116 &
  0.0978 \\

 \textbf{QD} &
%  MAE &
  0.5003 &
  0.4150 &
  0.3945 &
  0.3688 &
  0.6689 &
  0.4456 \\

 \textbf{RQspline} &
%  MAE &
  0.0917 &
  0.0622 &
  0.0479 &
  0.0485 &
  0.2153 &
  \textbf{0.0912} \\

\textbf{p-sline}   &
%  MAE &
  0.0778 &
  0.0570 &
  0.0444 &
  0.0453 &
  --- &
  0.0966 \\

\textbf{c-spline}  &
%  MAE &
  0.0806 &
  0.0543 &
  0.0430 &
  0.0447 &
  0.2002 &
 0.0947 \\

\rowcolor{Gray} \nsschainxy &
%  MAE &
  0.0787 &
  0.0588 &
  0.0430 &
  0.0448 &
  0.2052 &
  0.0962 \\

\rowcolor{Gray} \nsschainalphay  &
%  MAE &
  0.0846 &
  0.0568 &
  0.0417 &
  0.0448 &
  0.2067 &
  0.0976 \\

\rowcolor{Gray} \nsssum &
%  MAE &
  \textbf{0.0709} &
  \textbf{0.0512} &
  \textbf{0.0414} &
  \textbf{0.0442} &
  \textbf{0.1949} &
  0.0957 \\ \hline

  \textbf{Gain percentage} &
%  MAE &
   $12.0\%$ &
   $17.7\%$ &
   $3.7\%$ &
   $1.1\%$ &
   $2.6\%$
   &
  - \\
 \Xhline{2\arrayrulewidth} 
 % \Xhline{3\arrayrulewidth}
\end{tabular}%
}
\caption{\textbf{Mean Absolute Error (MAE) on UCI benchmarks.}  Test performance of the proposed method (NSS) and existing methods on UCI benchmarks. We use the $50$th quantile estimator as our estimates. The dash indicates unavailability. The shaded area is the proposed methods. Bold is the top one. Lower is better. \textbf{Gaussian}: Gaussian kernel;  \textbf{QD} is quantity-driven methods proposed in  \citep{pearce2018high}; RQ spline proposed in \citep{durkan2019neural}; 
c-spline proposed in \cite{gasthaus2019probabilistic}. \textbf{Boston, Concrete, Power} is short for Boston Housing, Concrete Strength, Power Plant. Gain percentage is computed as (best nss - best baseline)/best baseline. }
\label{tab:uci-measure}
\end{table*}

Once we select the operators and splines% (\secref{sec:methods})
, the parameters of the splines are trained in an end-to-end way by optimizing CRPS (\eqref{eq:crps}). 
Specifically, during training, we fit parameters by optimizing over with the empirical mean of CRPS over $N$ data points:
\begin{equation}
\theta^* = \arg\min_{\theta} 1/N \sum_{i=1}^N \mathbb{E}_y[\text{CRPS}(y, q_{\theta}(X_i, \alpha))].
\end{equation}
%The advantage of optimizing CRPS compared to quantile loss is that CRPS takes into account of all quantile levels, not only a single quantile. Therefore, the learned quantile function would be effective for any quantile target at inference. 
%\TODO{[Add connection paragraph] clarify training in the framework}
Algorithm \ref{alg:training} overviews the training of NSS for spline parameter selection. 
%The loss of the training model is CRPS (\eqref{eq:crps}), which involves integration over all quantile levels $\alpha$. 
Because of the form of the transformations, the analytical solution of CRPS integration is intractable. Thus, we use a Monte Carlo estimation for the CRPS loss. In particular, we sample $m$ number of $\alpha$ values from the range of $[0, 1]$ and average them for the corresponding pinball loss. %to obtain Monte Carlo estimator of the CRPS loss. Here we we empirical estimation of CRPS 
%\vspace{-2em}

%\begin{wrapfigure}{R}{0.45\textwidth}
    \begin{minipage}{0.45\textwidth}
    % \vskip -.3 in
    \removelatexerror% Nullify \@latex@error
\begin{algorithm}[H] %[ht]

\caption{Training with CRPS}\label{alg:training}
%\textbf{Operators} = \{+, $\times$, Scale, Chain, ...\} \\ 
%\textbf{Splines} = \{c-spline, p-spline, Gaussian, Causcy ...\} \\
%\KwData{Quantile level $\alpha \in [0, 1]$, $N$ data points $\{X \in \mathbb{R}^{d}, y \in \mathbb{R}^1\}_N, d \geq 1$. Depth $k$}
\KwData{$N$ data points $\{X_i \in \mathbb{R}^d, y_i \in \mathbb{R}^1\}_{i=1}^N$, $m$ quantile levels, $T$ transformation, which takes selected splines $S_{\text{select}}$ and selected operators $O_{\text{select}}$ from NSS. $lr$ is learning rate. } %The number of training epochs. }
%\KwInput{$n \geq 0$}
%\KwResult{$p(y|X)$ and $F^{-1}_{y|X}(\alpha)$}
\KwResult{Neural network weights $\theta$}
%$y \gets 1$\;
%$X \gets x$\;
$e \gets 1$\;
\While{$e \leq Nepoch$}{
%   \eIf{$N$ is even}{
%     \textbf{Normalize} $y$ to $[0, 1]$    %\Comment[r]*{Sigmoid or Scale^*}
%     $\alpha \gets y$\;
%   % $N \gets \frac{N}{2}$
%   }{\If{$N$ is odd}{
%       $y \gets y \times X$\;
%       $N \gets N - 1$\;
%     }
    
%   }
    
   f = \text{Transform}($S_{\text{select}}$, $O_{\text{select}}$)  %\Comment*[r]{Secified by NSS}  \;
   $\ell \gets 0$ \;
   %\For{$\alpha$ \text{ in linspace(}0, 1, num=$m$)}{  
\For{$\alpha$ \text{in [0,} $\frac{1}{m}$, $\frac{2}{m}$, ..$1$]}{  
   $y^{pred}_{\alpha}$ =  f$_{\theta}$($X$, $\alpha$) \;
   $ \ell \gets \ell$ + \text{pinball\_loss} ($y^{pred}_{\alpha}$, y, $\alpha$)
   }
   CRPS = $\ell / m$  \;
   %$\theta^* = \arg\min_{\theta}$ CRPS \;
   $\theta  \gets \theta  - lr \cdot \nabla_{\theta}$ CRPS \;
   $e \gets e + 1$\;
 }
\end{algorithm}
 \end{minipage} %{0.5\textwidth}
%\vskip -.15 in
%\vskip -.4 in
% \end{wrapfigure} %{L}{0.5\textwidth}

%% file: Experiments.tex
%\section{Experiments} \label{sec:experiments}
%\vspace{-.6em}
\subsection{Comparison methods}
%\vspace{-.1em}
%\paragraph{ Existing work:   }%\\

\textbf{QD} \citep{pearce2018high} generates prediction intervals (PIs) for estimating uncertainty for regression tasks with the assumption that high-quality PIs should be as narrow as possible. %In this paper we derived a loss function for the output of PIs based on the assumption that high-quality PIs should be as
%narrow as possible subject to a given coverage proportion. 
\textbf{Deep Quantile Aggregation} \citep{kim2021deep} proposes weighted ensembling strategies where aggregation weights vary over both
individual models and feature values plus (pairs of) quantile levels. The monotonization
layer in the network is applied to avoid crossing of quantile estimates. %The weights are generated by deep neural networks that adaptively combine arbitrary quantile estimates in a non-crossing manner, enforced through 
\textbf{RQspline} \citep{durkan2019neural} proposes a fully-differentiable module based on monotonic rational-quadratic splines,
which enhances the flexibility of coupling and autoregressive transforms while
retaining analytic invertibility.
\textbf{Global-Coarse} \citep{ratcliff1979group} provides an analysis of distribution statistics of group reaction time distributions.
\textbf{MLE (NB)} and \textbf{Mix. MLE} are Negative Binomial and 
 mixture likelihood based methods \citep{awasthi2021benefits}. %\TODO{Add ref}
\textbf{C-spline} is proposed in \citep{gasthaus2019probabilistic}, where C-spline is used as the quantile function in time-series forecasting.
%\vspace{-1em}

\subsection{Metrics}
%\paragraph{Metrics:  }
%\vspace{-.4em}
%\textbf{Metrics. }\\
For point predictions, we focus on the following metrics:
%\begin{itemize}[topsep=0pt, leftmargin=*]
{Mean absolute error (MAE)}:
$\frac{1}{n} \sum_{t=1}^n |T_t - P_t|$ where $T_t$ and $P_t$ are true and predicted value;
{Mean Absolute Percentage Error (MAPE)}:
$\frac{1}{n} \sum_{t=1}^n |\frac{T_t - P_t}{T_t}|$. %where $T_t$ and $P_t$ are true and predicted value.;
{Weighted Average Percentage Error (WAPE)}:
$\frac{\sum_{t=1}^n|T_t - P_t|}{\sum_{t=1}^n |T_t|}$; and
{Root Mean Square Error (RMSE)}: 
$\sqrt{\frac{\sum_t^N(T_t - P_t)^2}{n}}$.
%\end{itemize}
For quantile predictions, we use the Pinball Loss (\eqref{eq:pinballloss}), with $50\%$-th, \textbf{Q50}; $90\%$-th, \textbf{Q90}; and  $10\%$-th \textbf{Q10} quantiles. 
%\vspace{-.4em}
%\textbf{MAE}: Mean Absolute Error. 
%\textbf{MAPE}: Mean Absolute Percentage Error,  which is a measure of the accuracy of forecast system as a percentage. It is calculated as the average absolute percent error for each time period minus actual values divided by actual values. $$\frac{1}{n} \sum_{t=1}^n |\frac{T_t - P_t}{T_t}|$$ where $T_t$ and $P_t$ are true and predicted value. \\
%\textbf{WAPE}: Weighted Average Percentage Error, which is also referred as MAD/Mean ratio. 
%$$ \frac{\sum_{t=1}^n|T_t - P_t|}{\sum_{t=1}^n |T_t|}$$
%\textbf{RMSE}: Root Mean Square Error; 
%$$\sqrt{\frac{\sum_t^N(T_t - P_t)^2}{n}}$$
%\textbf{Pinball loss}: See \eqref{eq:pinballloss}. 
%\textbf{Q50}: pinball loss of $50\%$-th quantile .  
%\textbf{Q90}: pinball loss of $90\%$-th quantile. 
%\textbf{Q10}: pinball loss of 
%\textbf{Tuning}: 

\subsection{Training}
For simplicity, the proposed NSS methods use depth-2 splines, which contain \{(c-spline, p-spline), (c-spline, p-spline), (c-spline, c-spline), (p-spline, p-spline)\}. \nsssum is tuned with $\lambda$ in the range of $[0.1, 0.2, 0.3, 0.4, 0.5, 0.6, 0.7, 0.8, 0.9]$. 
\nsschain %is used with combination of p-spline and c-spline. 
normalizing of $y$ in $\alpha$ chaining can be achieved by applying sigmoid layer or scaling by max value. As splines are monotonically-increasing functions, the spline value $y$ with $\alpha=0$ is the minimum value of $y$ and $\alpha=1$ yields the maximum value of $y$. {\it Scale} is $y_{scale} = \frac{y - y_{min}}{y_{max} - y_{min}}$. We use a batch size=128 and a learning rate of $0.005$ for $100$ epochs.

%% file: Results.tex
%\section{Results} \label{sec:results}
%\vspace{-0.8em}

%\TODO{Add experiments details and dataset definitions. Descriptions of existing work}

%\nsschainxy, \nsschainalphay \nsssum
% \usepackage{graphicx}
\begin{table*}[h]
\centering
\resizebox{0.73\textwidth}{!}{\begin{tabular}{l|cccccc}
%\hline
 \Xhline{3\arrayrulewidth} 
\textbf{Methods} & \textbf{Boston} & \textbf{Concrete} & \textbf{kin8nm} & \textbf{Power} & \textbf{Protein} & \textbf{Wine} \\ 
\Xhline{2\arrayrulewidth}  %\hline
\textbf{Gaussian}             & 0.0276 & 0.0203 & 0.0171 & 0.0158 & 0.0725 & 0.0357 \\
\textbf{Global-Coarse$^*$}     & 0.0745 & 0.0596 & 0.0681 & 0.0473 & 0.1321 & ---    \\
\textbf{Deep Quantile Aggregation$^*$} & 0.0754 & 0.0541 & 0.0684 & 0.0441 & 0.1253 & ---    \\
\textbf{QD}                   & 0.1212 & 0.1076 & 0.1004 & 0.0972 & 0.1547 & 0.1164 \\
\textbf{RQspline}             & 0.0458 & 0.0418 & 0.0203 & 0.0189 & 0.0863 & 0.0424 \\
\textbf{p-sline}               & 0.0308 & 0.0211 & 0.016  & 0.0160  & ---    & 0.0358 \\
\textbf{c-spline}              & 0.0312 & 0.0198 & 0.0157 & 0.0159 & 0.0688 & \textbf{0.0351} \\
%\textbf{nas\_sum}             & 0.0265 & 0.0191 & 0.0152 & 0.0157 & 0.0674 & 0.0357 \\
\rowcolor{Gray} \nsschainxy  & 0.0311 & 0.0216 & 0.0165 & 0.0162 & 0.0707 & 0.0358 \\
\rowcolor{Gray} \nsschainalphay  & 0.0322 & 0.0208 & \textbf{0.0151} & 0.0159 & 0.0726 & 0.0363 \\ 
\rowcolor{Gray} \nsssum    & \textbf{0.0265} & \textbf{0.0191} & 0.0152 & \textbf{0.0157} & \textbf{0.0674} & 0.0357 \\ \hline

  \textbf{Gain percentage} &
%  MAE &
   $4.0\%$ &
   $3.5\%$ &
   $3.8\%$ &
   $0.6\%$ &
   $7.0\%$
    &
   -%1.7\% \\ 
   \\
\Xhline{3\arrayrulewidth} 
\end{tabular}%
}
\caption{\textbf{Average pinball loss on UCI benchmarks}. The test pinball loss (the lower, the better) is over 99 quantile levels, $\alpha=\{0.01, 0.02, ... 0.99\}$. The compared methods are Global-Coarse proposed in \citep{ratcliff1979group}; QD \citep{pearce2018high};
Deep Quantile Aggregation (DQA) \citep{kim2021deep};
RQspline \citep{durkan2019neural}; $*$ indicates entries are from \citep{kim2021deep} (under the same experiment setup).}
%\vskip -.1inch
\label{tab:ave-pinball loss}
\end{table*}
% \usepackage{graphicx}
\begin{table*}[ht]
\centering
\resizebox{0.7\textwidth}{!}{\begin{tabular}{l|cccccc}
 \Xhline{2\arrayrulewidth} 
\textbf{Methods} & \textbf{MAPE} & \textbf{WAPE} & \textbf{RMSE} &  \textbf{Q50} & \textbf{Q90} & \textbf{Q10} \\ \Xhline{2\arrayrulewidth} 
\textbf{MLE (NB)}                   & \textbf{0.44434} & 0.27240  & 7.70958 & 0.27240  & 0.10907 & 0.15275 \\
\textbf{Mix MLE}                    & 0.44839 & 0.26838 & 7.22556 & 0.26838 & 0.10293 & 0.14508 \\
\textbf{c-spline}                   & 0.44672 & 0.26635 & 7.06332 & 0.26635 & \textbf{0.10238} & 0.14241 \\
\textbf{p-spline}                   & 0.44912 & 0.26834 & 7.14643 & 0.26834 & 0.10343 & 0.14333 \\
\rowcolor{Gray} 
\textbf{\nsssum}                    & 0.44501 & 0.26545 & 6.96697 & 0.26545 &  \textbf{0.10238}  & 0.14266 \\
\rowcolor{Gray} 
\textbf{\nsschain}                  & 0.44883 &  \textbf{0.26420} &  \textbf{6.91726} &  \textbf{0.26420}  & 0.10243 &  \textbf{0.14149} \\
 \Xhline{2\arrayrulewidth} 
\end{tabular}%
%\vskip -1 inch
}
\caption{\textbf{Performance comparisons for time series forecasting on M5}. Different evaluation metrics are included in this table for M5. Detailed descriptions of the metrics are in \secref{sec:experiments}. $Q_k$ indicates the pinball loss of $k$-th quantile. e.g. $Q50$ is the pinball loss of $50$th quantile.  Lower is better.}
\label{tab:time-series}
%\vskip -0.2in
\vspace{-.5em}
\end{table*}

To demonstrate the effectiveness of proposed methods, we conduct experiments on synthetic, real-world tabular regression, and time series forecasting datasets. 
%\vspace{-1.3em}
\subsection{Synthetic data} \label{sec:Synthetic}
\begin{figure}[t]
%\vskip -0.3in
\centering
\includegraphics[width=\columnwidth]{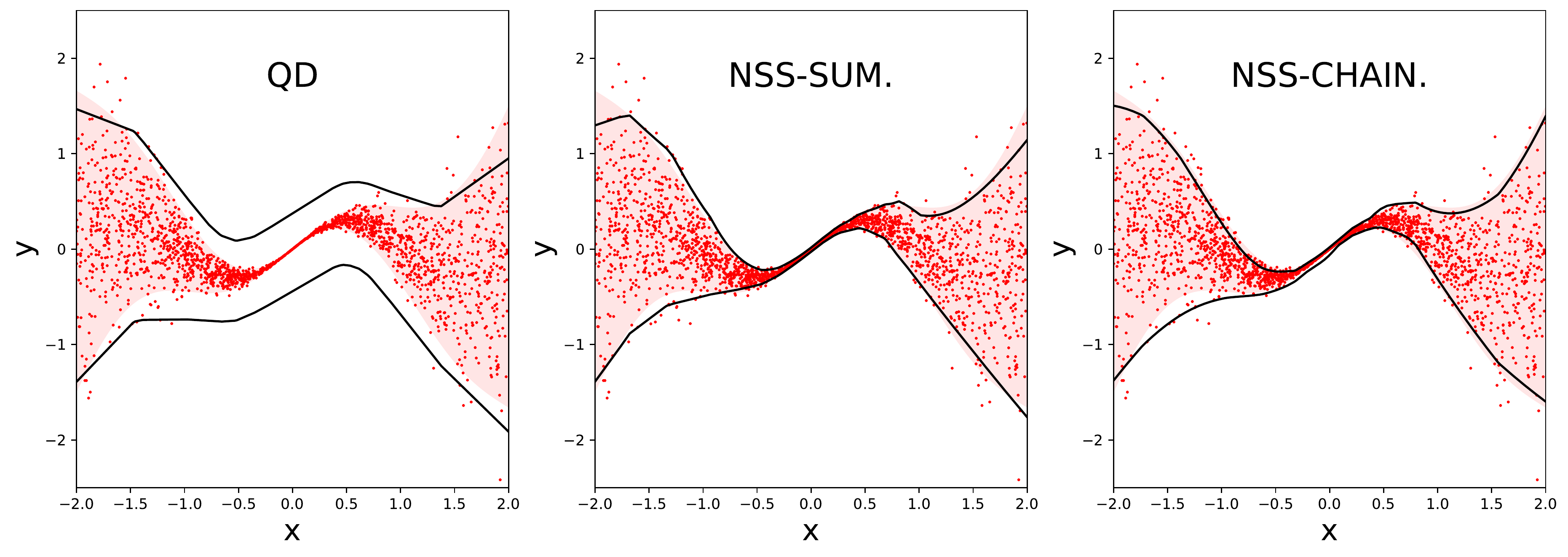}
\caption{\textbf{NSS on Synthetic data}. We compare the performance of proposed NSS against existing methods $QD$ \citep{pearce2018high}. The red dots are observed data points, shaded red area is the ground truth $2.5\%$ and $97.5\%$ quantile levels, and the dark black lines are the predicted $2.5\%$ and $97.5\%$ quantile levels.}\label{fig:synthetic data}
\vskip -0.07in
\end{figure}

\begin{figure}[t]
%\vskip -0.1in
\centering
\includegraphics[width=\columnwidth]{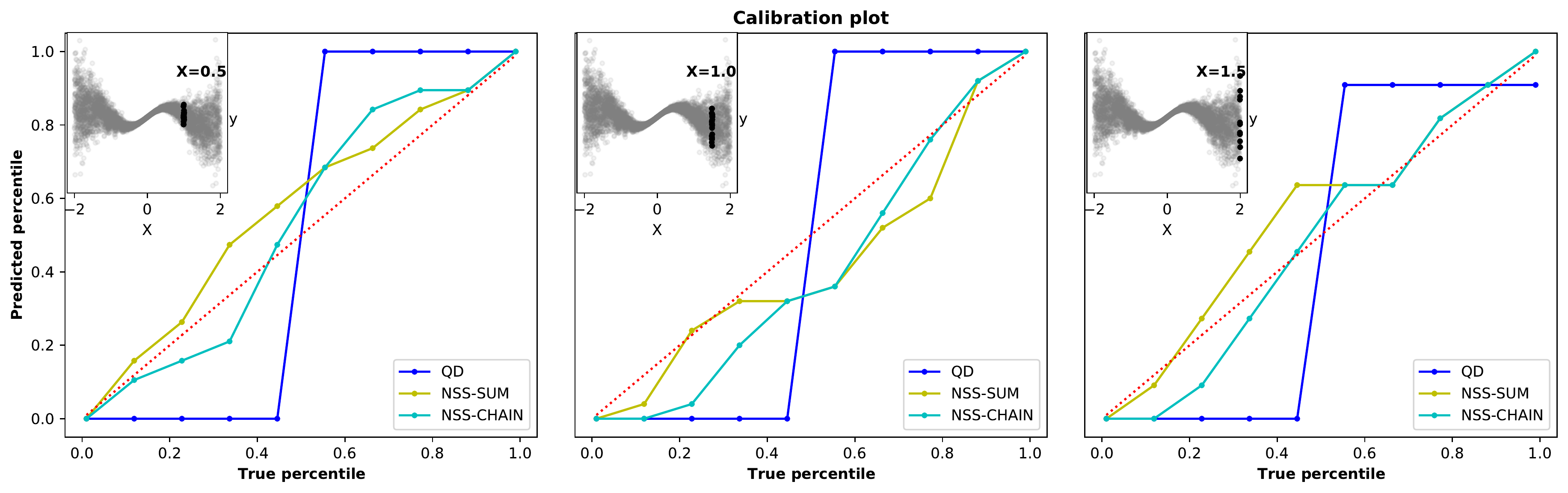}
\caption{ \textbf{Calibration plots}. Predicted vs. ground truth percentiles at condition levels: $X$=0.5, 1.0 and 1.5. The perfect calibration would correspond to the diagonal (red dotted) line.}\label{fig:calibration}
%\vskip -0.15in
\vspace{-1em}
\end{figure}

%\TODO{Add figures and results}
\textbf{Dataset}. We generate 2000 data points ($X \in \mathbb{R}^1$ and $y \in \mathbb{R}^1$), where $X$ is in the range of $[-2, 2]$ and $y$ has Gaussian distribution $y \sim \mathcal{N}(0.3 \sin(3x), 0.2 x^2)$, where $\sin$ is the sinusodial function. 
%The variance scales quadratically with $x$. 
We construct the validation and test sets to come from the same distribution. 
%We first evaluate on Sin-Gaussian datasets, where $y$ is Gaussian distributed with non-fixed variance. 
Unlike real-world data, the synthetic data would have known quantile levels, that can be used for evaluating the accuracy of quantile estimates. We make the task more challenging by setting a data-dependent variance for the Gaussian noise to evaluate the ability of learning condition-specific quantile values. \figref{fig:synthetic data} shows that the proposed NSS-chain and NSS-sum can capture the true underlying quantiles, whereas QD \citep{pearce2018high} struggles on the varying variance locations (e.g. around $x = 0$). The upper and lower black lines are the predicted 2.5\%-th and 97.5\%-th quantiles for the observed data (e.g. red dots), shown along with the ground truth quantiles (e.g. shaded red area). The results indicate that more expressive NSS transformations are superior in more challenging scenarios, where true data points are distributed differently (e.g., distributions depend on the value of the inputs").  
\figref{fig:calibration} shows the calibration plot of the predicted vs. true distributions at different quantile levels. Here, we show the true percentile $p$ as the fraction of data in the dataset such that the $p$ percentile of the predictive distribution is larger than the ground truth data. The perfect prediction would be the diagonal line. \figref{fig:calibration} indicates that the proposed methods NSS-sum and NSS-chain can capture the proposed true distribution at various levels by close to the red line, whereas QD does not fit as well. 

\subsection{Real-world tabular regression} \label{sec:uci}
%We present the UCI datasets results on \tabref{tab:uci-measure} and pinball loss \tabref{tab:ave-pinball loss}

We use UCI benchmarks  \citep{asuncion2007uci} that contain tabular data from diverse domains (e.g. real estate and physics). Following \citep{salem2020prediction}, the datasets are normalized with z-score standardization.
%(subtract mean and divide by standard deviation). 

We evaluate the accuracy for both point predictions and quantiles. As the point predictions, we use the $50$th quantile estimator as our estimates. \tabref{tab:uci-measure} shows that the proposed NSS methods outperform the other existing methods on most datasets in mean absolute error (MAE). In mean square error (MSE), the results are provided in Appendix \tabref{tab:uci-measure MSE}. We observe that the NSS-sum performs better than NSS-chain. For quantile metrics, we use the pinball loss (\eqref{eq:pinballloss}) over 100 quantile levels $\alpha=\{0.01, 0.02, ... 0.99\}$ in \tabref{tab:ave-pinball loss}. The results indicate that NSS consistently outperforms other alternatives across different UCI benchmarks. In pinball loss, NSS-sum performs better than NSS-chain. We attribute the superiority of NSS-sum for regression to make balance between different transformation, which is helpful in explaining the variance in the data. 
%\vspace{-.9em}

\subsection{Retail demand forecasting} \label{sec:m5}
%\TODO{add Chun-liang's results about M5, and run NAS results on M5. }
%\TODO{M5 description?}
For time series forecasting, we focus on the M5 dataset, which contains time-varying sales data for retail goods, along with other relevant covariates like price, promotions, day of the week, special events etc. It represents  an important real-world scenario, where the accurate estimation of the output distribution is crucial, as retailers use them to optimize %the supply-chain planning,
prices or promotions. \\

The time series forecasting experiments are conducted by performing one-step ahead prediction, yielding predictions in an autoregressive way. \tabref{tab:time-series} shows the results of our method compared to other alternatives. We observe consistent outperformance of NSS in various forecasting evaluation metrics. Different from regression tasks, we observe that NSS-chain is better than NSS-sum, indicating its benefit in capturing time-dependent relationship.    \\

\textbf{Remarks} on NSS-sum vs NSS-chain. The results show that NSS-sum is superior on regression, while NSS-chain has advantages on time series forecasting. The observations may indicate NSS-sum is suitable for more constrained tasks (e.g. regression, one time step time series-forecasting), where being moderately expressive would suffice. NSS-sum is also more robust and easier to train. On the other hand, NSS-chain may be more expressive, which is beneficial to fit tasks requires more complex distributions at different time steps of the time series, but for individual step NSS-chain is not as accurate as NSS-sum in fitting the distribution. 

%\vspace*{-\baselineskip}
%\section{Discussion} \TODO{}
%\section{Future work}
%Additonaly, NSS can be integrated into previous work in modeling the uncertainty by serving as their quantile function.  The majority previous efforts on modeling uncertainty For example, in time series forecasting, one can incorporate the proposed quantile function to various sequential neural network state space models \cite{wen2017multi, salinas2020deepar}.The proposed quantile function can also be used as a basis for ensembling of quantile models \citep{pearce2018high, lakshminarayanan2016simple,  kim2021deep}, or priors for models such as Bayesian neural networks \citep{hernandez2015probabilistic, blundell2015weight, kendall2017uncertainties, teye2018bayesian}. 

\section{Conclusion}
We propose a novel approach for modeling uncertainty. %, by utilizing learnable splines to transform quantile space. 
The proposed {\it Neural Spline Search (NSS)} method employs a series of monotonic spline regression transformations, guided by symbolic operators. 
We demonstrate the effectiveness of NSS for superior modeling of output distributions, on both synthetic and real-world datasets. 
We leave the extensions to different operators and splines, including parametric distribution transformations to future work.

\clearpage

%% file: Appendix.tex
\section{Appendix}
\textbf{Mean square error for UCI dataset}
\begin{table*}[ht]
\centering
\resizebox{\textwidth}{!}{%
\begin{tabular}{l|c|c|c|c|c|c}
%\hline
\Xhline{4\arrayrulewidth}
\textbf{Methods} &
 % \textbf{Measures} &
  \textbf{Bost House}&
  \textbf{Concr Stren} &
  \textbf{kin8nm} &
 \textbf{Power plant} &
  \textbf{Protein} &
  \textbf{Wine} \\
  %\hline
\Xhline{3\arrayrulewidth}
  \textbf{Gaussian} &
%   MAE &
%   0.0754 &
%   0.0564 &
%   0.048 &
%   0.0449 &
%   0.2116 &
%   0.0978 \\
% \multirow{-2}{*}{\textbf{Gaussian}} &
 %MSE &
 \textbf{0.0105} &
0.0054 &
0.0042 &
 \textbf{0.0032} &
0.0648 &
 0.0164  \\  %\Xhline{2\arrayrulewidth}
 \textbf{\citep{salem2020prediction}$^*$}  &
%   MAE &
%   --- &
%   --- &
%   --- &
%   --- &
%   --- &
%   --- \\
% \multirow{-2}{*}{\textbf{Salem et al {[}1{]} *}} &
% MSE &
0.1120 &
0.0560 &
0.0600 &
0.0420 &
0.3100 &
0.5970 \\  %\Xhline{2\arrayrulewidth}
 \textbf{QD}  &
%   MAE &
%   0.5003 &
%   0.415 &
%   0.3945 &
%   0.3688 &
%   0.6689 &
%   0.4456 \\
% \multirow{-2}{*}{\textbf{qd {[}2{]}} }&
% MSE &
0.2705 &
0.1839 &
0.1613 &
0.1393 &
0.5277 &
0.2164 \\  %\Xhline{2\arrayrulewidth}
 \textbf{RQspline}  &
%   MAE &
%   0.0917 &
%   0.0622 &
%   0.0479 &
%   0.0485 &
%   0.2153 &
%   \textbf{0.0912} \\
% \multirow{-2}{*}{\textbf{RQspline}} &
%  MSE &
0.0255 &
0.0070 &
0.0040 &
0.0037 &
0.0809 &
0.0195 \\  %\Xhline{2\arrayrulewidth}
\textbf{p-sline}   &
%   MAE &
%   0.0778 &
%   0.057 &
%   0.0444 &
%   0.0453 &
%   --- &
%   0.0966 \\
% \multirow{-2}{*}{\textbf{psline}} &
%MSE &
0.0136 &
0.0058 &
0.0032 &
0.0032 &
 --- &
 0.0162 \\  %\Xhline{2\arrayrulewidth}
\textbf{c-spline}  &
%   MAE &
%   0.0806 &
%   0.0543 &
%   0.043 &
%   0.0447 &
%   0.2002 &
%  0.0947 \\
% \multirow{-2}{*}{\textbf{cspline}} &
% MSE &
 0.0162 &
 0.0050 &
 0.0031 &
 0.0032 &
 0.0757 &
 \textbf{0.0159} \\ %\hline
% &
%   MAE &
%   \textbf{0.0709} &
%   \textbf{0.0512} &
%   \textbf{0.0414} &
%   \textbf{0.0442} &
%   \textbf{0.1949} &
%   0.0957 \\
% \multirow{-2}{*}{\textbf{nas\_sum}} &
%   \colorgray MSE &
%   \colorgray 0.0112 &
%   \colorgray \textbf{0.0046} &
%   \colorgray \textbf{0.0029} &
%   \colorgray \textbf{0.0032} &
%   \colorgray 0.0711 &
%   \colorgray 0.016 \\  \Xhline{2\arrayrulewidth}
\rowcolor{Gray}\nsschainxy   &
%   MAE &
%   0.0787 &
%   0.0588 &
%   0.043 &
%   0.0448 &
%   0.2052 &
%   0.0962 \\
% \multirow{-2}{*}{\textbf{nas\_chain (xy loop)}} &
% \text{MSE} &
 0.0128 &
0.0056 &
0.0031 &
\textbf{0.0032} &
0.0751 &
0.0164 \\ %\Xhline{2\arrayrulewidth}
\rowcolor{Gray}\nsschainalphay  &
%   MAE &
%   0.0846 &
%   0.0568 &
%   0.0417 &
%   0.0448 &
%   0.2067 &
%   0.0976 \\
% \multirow{-2}{*}{\textbf{nas\_chain (qy loop)} }&
% \text{MSE} &
0.0184 &
0.0058 &
\textbf{0.0029} &
\textbf{0.0032} &
0.0760 &
0.0169 \\ 
  %\hline

\rowcolor{Gray}\textbf{\nsssum}  &
%   MAE &
%   \textbf{0.0709} &
%   \textbf{0.0512} &
%   \textbf{0.0414} &
%   \textbf{0.0442} &
%   \textbf{0.1949} &
%   0.0957 \\
% \multirow{-2}{*}{\textbf{nas\_sum}} &
% MSE &
0.0112 &
\textbf{0.0046} &
 \textbf{0.0029} &
\textbf{0.0032} &
0.0711 &
 0.0160 \\  %\Xhline{2\arrayrulewidth} 
  
  \Xhline{4\arrayrulewidth}
\end{tabular}%
}
\caption{Mean Square Error of UCI datasets}
\label{tab:uci-measure MSE}
\end{table*}